\documentclass[10pt,twocolumn,letterpaper]{article}

\usepackage{cvpr}              
\usepackage{comment}

\definecolor{cvprblue}{rgb}{0.21,0.49,0.74}
\usepackage[pagebackref,breaklinks,colorlinks,allcolors=cvprblue]{hyperref}

\def\paperID{36} 
\def\confName{CVPR}
\def\confYear{2025}

\title{Analyzing and Fine-Tuning Whisper Models for Multilingual Pilot Speech Transcription in the Cockpit}

\author{Kartheek Kumar Reddy Nareddy\\
Institute of Data Science\\
German Aerospace Center\\
{\tt\small kartheek.nareddy@dlr.de}
\and
Sarah Ternus\\
Institute of Flight Guidance\\
German Aerospace Center\\
{\tt\small sarah.ternus@dlr.de}
\and
Julia Niebling\\
Institute of Data Science\\
German Aerospace Center\\
{\tt\small Julia.Niebling@dlr.de}
}

\begin{document}
\maketitle

\begin{abstract}
The developments in transformer encoder-decoder architectures have led to significant breakthroughs in machine translation, Automatic Speech Recognition (ASR), and instruction-based chat machines, among other applications. The pre-trained models were trained on vast amounts of generic data over a few epochs (fewer than five in most cases), resulting in their strong generalization capabilities. Nevertheless, the performance of these models does suffer when applied to niche domains like transcribing pilot speech in the cockpit, which involves a lot of specific vocabulary and multilingual conversations. This paper investigates and improves the transcription accuracy of cockpit conversations with Whisper models. We have collected around 85 minutes of cockpit simulator recordings and 130 minutes of interview recordings with pilots and manually labeled them. The speakers are middle aged men speaking both German and English. To improve the accuracy of transcriptions, we propose multiple normalization schemes to refine the transcripts and improve Word Error Rate (WER). We then employ fine-tuning to enhance ASR performance, utilizing performance-efficient fine-tuning with Low-Rank Adaptation (LoRA).  Hereby, WER decreased from 68.49 \% (pretrained whisper Large model without normalization baseline) to 26.26\% (finetuned whisper Large model with the proposed normalization scheme).
\end{abstract}
    
\section{Introduction}
\label{sec:intro}
Automatic Speech Recognition (ASR), transforming audio signals into text, plays a key role in natural language processing~\cite{benzeghiba2007automatic}. The diversity in speech signals with variations like gender, accent, pace, external noise, etc. makes ASR a challenging problem \cite{benzeghiba2007automatic}. ASR has found applications in automatic call handling \cite{das2002application}, and personalized AI assistants 
\cite{matarneh2017speech}. Conventional ASR systems rely on a pipeline of components, including acoustic feature extraction, acoustic and language modeling, and decoding via Bayes' decision rule~
\cite{bayes1763essay}.
With the advent of deep learning, both acoustic and language modeling have been revolutionized
~\cite{bengio2003neural},
ultimately leading to end-to-end models
~\cite{collobert2011natural}.\\
\indent Publicly available datasets like LibriSpeech~\cite{panayotov2015librispeech}, Common Voice~\cite{ardila2019common}, and SpeechStew~\cite{chan2021speechstew} contributed towards training and testing newly upcoming ASR models. However, the increasing size of neural networks has outpaced the size of these labeled datasets, often resulting in overfitting and poor generalization~\cite{geirhos2020shortcut}. This challenge has motivated the creation of large-scale unlabeled or weakly labeled datasets, such as BigSSL (1 million hours)~\cite{zhang2022bigssl}, GigaSpeech~\cite{chen2021gigaspeech}, and People's Speech~\cite{galvez2021people}. Among contemporary ASR models, OpenAI's Whisper stands out for its large-scale, weakly supervised training across 680,000 hours of multilingual data, incorporating both supervised and unsupervised techniques to achieve broad generalizability across diverse domains and languages~\cite{van2024whisper}.\\
\indent While Whisper and other transformer-based models (e.g., Wav2Vec \cite{baevski2020wav2vec2}, SpeechStew\cite{chan2021speechstew}, DeepSpeech\cite{amodei2016deepspeech2}) perform impressively on general speech data, their accuracy can degrade in domain-specific contexts \cite{williams2023asrclassroom}. Fine-tuning, the process of adapting a pre-trained model to a specific task or dataset, has emerged as a powerful method to enhance ASR performance under such conditions \cite{liao2023prompt}. Fine-tuning has proven effective across diverse application areas, including healthcare and low-resource languages. For example, adapting Whisper for Nepali speech led to substantial reductions in Word Error Rate (WER), with improvements up to 36.2\% on the small model~\cite{rijal2023nepaliwhisper}.\\ 
\indent In aviation, ASR has also been widely explored for Air Traffic Control (ATC) communication. Domain-specific ASR models like Whisper-ATC have achieved as low results as 1.17\% WER on ATCOSIM simulated data and up to 60\% improvement through regional fine-tuning~\cite{van2024whisper}. However, ASR for intra-cockpit communication between pilots and the fine-tuning of ASR models for that use-case remains relatively unexplored. Accurate transcription in this context can support human factors research, assess teamwork dynamics, and lay the groundwork for speech-driven cockpit automation systems. Studies in the past considered hidden markov models based transcription technologies aiming to transcribe cockpit conversations \cite{Schmidt2016fraunhofer, Papenfuss2023using}. Yet, the cockpit environment poses unique challenges, including overlapping speech, multilingual exchanges, high noise levels and a lot of use-case specific vocabulary. \\
\indent To tackle these challenges, we explore the fine-tuning of Whisper models for multilingual pilot communication in the cockpit. Thereby, we adopt the Hugging Face fine-tuning pipeline. Our contributions aim to give an overview over fine-tuning Whisper for this domain, analyze model performance across different Whisper models and scenarios, proposing new normalization schemes, and establish groundwork for future ASR applications in the cockpit.

\section{Methodology}
\subsection{Dataset}

The dataset consists of ~85 minutes of cockpit simulator recordings and ~130 minutes of pilot interviews. The recordings cover various cockpit communication scenarios, including checklists, briefings, and emergency procedures, and reflect typical cockpit vocabulary. The audio is multilingual, with a mix of German and English as commonly spoken by German pilots. All audio was converted to MP3, segmented into 30-second clips, and resampled to 16 kHz. Manual transcripts were created as ground-truth references. The speakers are middle-aged male pilots.

\subsection{Metrics}
To evaluate transcription accuracy, WER was used as the primary metric. WER was computed using the jiwer library\footnote{https://jitsi.github.io/jiwer/}, which provides a standardized implementation for text-based error measurement. WER is a common metric in speech recognition and is defined as, WER~=~$\displaystyle \frac{S + D + I}{N}$,   
where \( S \) represents the number of substitutions, \( D \) the number of deletions, \( I \) the number of insertions, and \( N \) the total number of words in the reference text. A lower WER indicates a more accurate transcription. 
\subsection{Transcript Normalization}
\label{subsec:transcript_normalize}
 In this paper, various normalization schemes are being compared. First, three normalization steps from Whisper were applied: Basic normalization, which includes case lowering and the removal of special characters; Number normalization, which converts numeric expressions into Arabic numerals; and the English normalizer, which combines text, number, and spelling normalization.\\
\indent In addition to these, we introduce Proposed I, a custom normalization function incorporating similar processing for numbers, spelling, and punctuation with additional functions for transforming the ICAO-alphabet into standard letters (e.g., "DELTA" into "D") , removing filler words, and normalizing compound words (e.g., ensuring "take-off," "takeoff," and "take off" are treated as equivalent). Lastly, we evaluated two combined approaches: Proposed II, which applies Proposed I first, followed by English normalizer, and Proposed III, which applies English normalizer first, followed by Proposed I. Throughout the combined normalization approaches, the spelling and number normalizers are solely taken from the Whisper English normalizer.
\subsection{Finetuning}
The fine-tuning step of this study was conducted using the audio files and transcriptions from Section 2.1. For fine-tuning, the dataset was divided into a training set consisting of 158 audio files and a test set containing 40 audio files. 
Furthermore, the HuggingFace Transformers Python package was utilized to handle the fine-tuning procedure. Labels were extracted using the Whisper tokenizer. The log-mel spectrogram was computed using the feature extractor and processed as features. The data collator was used to ensure that the length of the features matched that of the input tokens. We used the LoRA \cite{hu2021lora} fine-tuning method, with learnable parameters amounting to approximately 1\% of the model's total parameters. Fine-tuning of the Whisper models was performed using an NVIDIA Tesla V100. Hereby, multiple learning rates from \{1e-5, 1e-4, 1e-3\} were tested. 

\section{Results \& Discussion}
\subsection{Baseline Results}
\begin{figure}[t]
\centering
    \includegraphics[width=\linewidth]{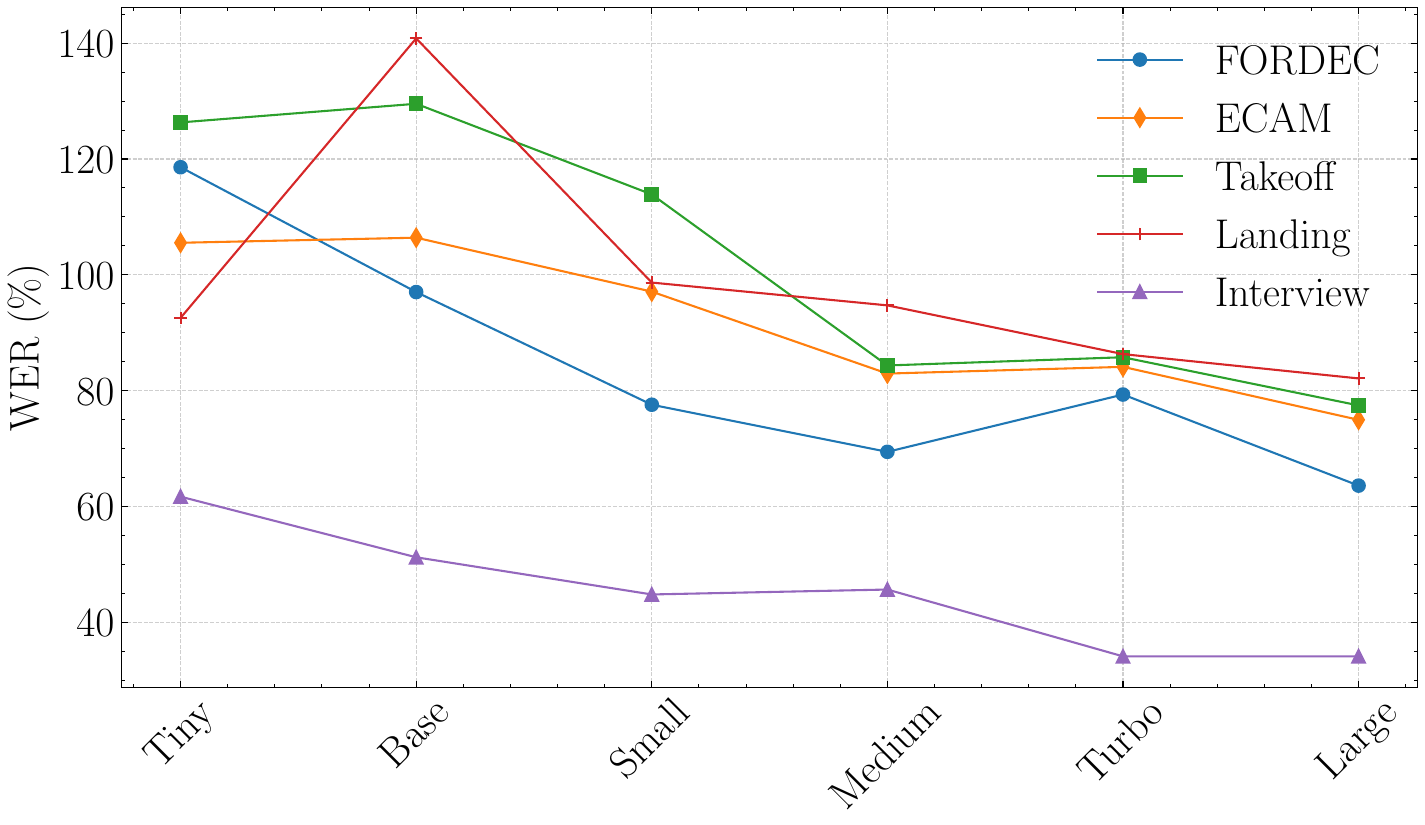}
    \vskip -0.05in
    \caption{Baseline Word Error Rate (WER) comparison across family of Whisper models for various pilot speech scenarios.}
    \label{fig:baseline_plot}
\end{figure}

We transcribed the audio files across the five scenarios using the family of multilingual whisper models. Then we computed the WER between predictions and the reference transcriptions. The results are shown in Figure~\ref{fig:baseline_plot}. Whisper Tiny and Base models have WERs exceeding 100\% in a few cases, indicating notable transcription errors. The Small and Medium Whisper models have considerable performance improvement over Tiny and Base, with WER in the 75-85\% range on average. Whisper Turbo has 73.92\% mean WER and Large-v3 (henceforth called as Large) has 66.44\% WER, indicating the necessity for fine-tuning and text normalization.
\subsection{Effect of Normalization}
Table~\ref{table:normalizers_baseline} shows a comparison of normalization schemes considered in this paper. The raw text predicted by the Whisper model is noted as no-norm(alization) text and has the highest WER. Basic text normalizer does have decent performance improvement over no-norm, while Number normalizer is falling behind the basic text normalizer. English normalizer on the other hand has best performance among the baseline normalizers considered. Three normalization schemes are presented in this paper, namely, Proposed I, II, and III. Among these, Proposed II and Proposed III have the lowest WERs for most of the Whisper models. This shows that normalizing ICAO-alphabets and removing filler words in combination with English normalizer results in the best performance.
\begin{table}[t]
\centering
\caption{WER (in \%) comparison of proposed normalization methods against baselines across family of Whisper models.}
\resizebox{\linewidth}{!}{
\begin{tabular}{|c|c|c|c|c|c|c|}
\hline
Normalizer  & Tiny & Base & Small & Medium & Turbo & Large \\ \hline 
No-norm  & 94.41 & 96.00 & 85.64 & 81.64 & 70.20 & 68.49 \\  
Basic  & 91.73 & 84.70 & 69.35 & 62.96 & 49.49 & 52.23 \\  
Number  & 89.13 & 88.70 & 77.41 & 70.84 & 62.18 & 59.76 \\   
English  & 88.37 & 84.58 & 69.16 & 62.43 & 48.88 & 52.08 \\ 
Proposed I  & 85.68 & 83.10 & 69.05 & 63.19 & 49.68 & 52.74 \\ 
Proposed II  & 88.41 & 84.25 & 69.87 & 62.60 & 48.69 & 52.00 \\ 
Proposed III  & 88.21 & 82.96 & 68.76 & 62.54 & 48.70 & 52.41 \\   \hline
\end{tabular}}
\label{table:normalizers_baseline}
\vskip -0.2in
\end{table}
\subsection{Effect of Fine-Tuning}
Fine-tuning the Whisper language models further resulted in improved transcription accuracy. The results of LoRA fine-tuning, with approximately 1\% of learnable parameters, on Whisper Large and Turbo models are given in Tables~\ref{table:lora_large}~and~\ref{table:lora_turbo}. The fine-tuning results of the remaining Whisper models and relevant Python scripts can be found in the supplementary material. The Whisper Turbo model, when fine-tuned with $6,553,600$ parameters, representing $0.8\%$ of its $815,431,680$ total parameters, achieved a WER reduction from 70.20\% to 61.82\% without any normalization. Similarly, the Whisper Large model showed a drop in WER from 68.49\% to 55.65\% with fine-tuning alone, depending on the learning rate. We experimented with different learning rates, and the optimal values varied between models. Whisper Turbo gave the best results at a learning rate of 1e-5, while Whisper Large models performed best with a 1e-3 learning rate.

\subsection{Combining Normalization and Fine-Tuning}
As shown in Section~3.2, normalization alone provides a notable improvement in transcription accuracy. However, an interesting finding is that its effect becomes even more pronounced when applied after fine-tuning. For example, the Whisper Large model had a WER of 68.49\% without normalization, which decreased to 52.00\% when the proposed II normalizer was applied to the pre-trained model. After fine-tuning (with a learning rate of 1e-3), the model’s WER without normalization was 55.65\%, and further decreased to 26.26\% when combined with the same normalization method. This demonstrates that while normalization improves performance on its own, its impact is more pronounced after the model has been fine-tuned. A similar trend was observed for the Whisper Turbo model, where the WER dropped from 70.20\% (pre-trained, no normalization) to 61.82\% after fine-tuning with a 1e-5 learning rate. When English normalization was applied, the WER further reduced to 28.01\%. These results suggest that combining normalization with fine-tuning can yield greater improvements than using either approach independently.

\begin{table}[t]
\centering
\caption{LoRA fine-tuning on Whisper Large model with various learning rates. The numbers indicate WER in \%.}
\resizebox{\linewidth}{!}{
\begin{tabular}{|c|c|c|c|c|}
\hline
Normalizer  & pre-trained & lr=1e-5 & lr=1e-4 & lr=1e-3  \\ \hline No-norm  & 68.49 & 58.83 & 64.36 & 55.65 \\  
Basic  & 52.23 & 27.96 & 37.09 & 27.37 \\ 
Number  & 59.76 & 46.10 & 48.30 & 50.08 \\ 
English  & 52.08 & 27.80 & 36.71 & 26.36 \\ 
Proposed I  & 52.74 & 32.72 & 37.35 & 38.37 \\ 
Proposed II  & 52.00 & 27.65 & 36.41 & 26.26 \\ 
Proposed III  & 52.41 & 28.24 & 36.60 & 27.00 \\    \hline
\end{tabular}}
\label{table:lora_large}
\end{table}
\begin{table}[t]
\centering
\caption{LoRA fine-tuning on Whisper Turbo model with various learning rates. The numbers indicate WER in \%.}
\resizebox{\linewidth}{!}{
\begin{tabular}{|c|c|c|c|c|}
\hline
Normalizer  & pre-trained & lr=1e-5 & lr=1e-4 & lr=1e-3  \\ \hline No-norm  & 70.20 & 61.82 & 64.67 & 65.06 \\  
Basic  & 49.49 & 28.18 & 29.04 & 31.02 \\  
Number  & 62.18 & 43.08 & 46.64 & 47.61 \\  
English  & 48.88 & 28.01 & 28.81 & 30.40 \\  
Proposed I  & 49.68 & 29.17 & 29.98 & 31.70 \\  
Proposed II  & 48.69 & 28.24 & 28.88 & 30.32 \\   
Proposed III  & 48.71 & 28.40 & 28.67 & 30.55 \\      \hline
\end{tabular}}
\label{table:lora_turbo}
\end{table}

\section{Future Work}
Though fine-tuning enhanced the ASR performance, the reported 26\% WER in Section 3.3 is not suitable for reliable deployment. One promising direction is to utilize prompting to provide context and aid in recognition of domain-specific vocabulary. Further, we aim to generate more data for improving the fine-tuning performance. The WER computation does play a crucial role in determining the suitability of the language model for transcription. A context-based WER computation that overlooks minor grammatical variations typical of spoken language could provide a more accurate reflection of ASR model performance. Therefore, further improvements in normalization, as well as methods to assess whether the transcribed content conveys the intended meaning, should be considered.  
\section{Conclusion}
In this paper the transcription of cockpit conversations using Whisper language models was explored. The audio files contain conversations between pilots in both German and English languages. The Whisper models transcribed the conversations with a high WER, which necessitates normalization and fine-tuning. Thereby, whisper normalization was utilized and own normalization schemes to normalize ICAO-alphabet, compound words and remove filler words were introduced, which resulted in better performance. Finetuning with Low-rank adaptation combined with normalization resulted in reduction of WER from 70.20\% (pre-trained Whisper Turbo model without normalization baseline) to 28.01\% (fine-tuned Whisper Turbo model with the Proposed II normalization scheme) on the test dataset.
The results emphasize the importance of domain adaptation for ASR models, particularly with technical vocabulary, multi-lingual speech, etc. Future work could include the exploration of prompting strategies, the creation of more training data, and more effective error computation approaches to further enhance the performance.
{
    \small
    \bibliographystyle{ieeenat_fullname}
    \bibliography{main}
}

\clearpage
\setcounter{page}{1}
\maketitlesupplementary


\section{Additional Results}
\subsection{Dataset Adaptation: Scenario Comparison}
In the supplementary results, we additionally compare transcription performance across four distinct operational scenarios: Takeoff briefings and checklists (10 scenarios, ~20 minutes), ECAM actions (11 scenarios, ~30 minutes), FORDEC decision-making procedures (3 scenarios, ~15 minutes), and landing briefings and checklists (5 scenarios, ~20 minutes). Additionally, a controlled interview scenario incorporating aviation-specific vocabulary was included for comparison (12 scenarios, ~130 minutes).

\subsection{Effect of Normalization}
A comparison of different normalization schemes is presented in Tables~\ref{table:normalizers_ecam}~to~\ref{table:normalizers_takeoff}.
The evaluation of various normalizers across a family of Whisper models on five distinct scenarios: ECAM, FORDEC, Interview, Landing, and Takeoff shows a consistent trends in performance improvements. Across all scenarios, the No-norm baseline exhibits the highest word error rate (WER), indicating that raw model outputs contain significant transcription errors. Among the baseline normalizers, the Basic and English approaches consistently outperform the Number normalizer, with notable reductions in WER. The proposed normalization techniques further refine these results, with Proposed II and Proposed III showing the most robust performance across different Whisper models. Larger models (Turbo and Large) tend to benefit more from normalization than smaller models (Tiny and Base), suggesting that model capacity influences the effectiveness of text normalization.\\
\indent In the ECAM and FORDEC scenarios, the Proposed II and Proposed III normalizers achieve the lowest WER for Medium, Turbo, and Large models. Specifically, in ECAM, Proposed II achieves a WER of 49.48 for Large, while in FORDEC, Proposed III achieves a WER of 43.09 for Large. The Interview scenario follows a similar trend, with Proposed II yielding the best results across most model sizes, achieving a WER of 23.75 \% for Large. The English normalizer performs comparably well, often ranking close to Proposed II. For Landing scenario, Proposed II achieves the lowest WER of 64.86 for Large, whereas for Takeoff, Proposed III yields the lowest WER of 44.89. Overall, the results emphasize the importance of selecting appropriate normalization strategies to enhance ASR accuracy, particularly in specialized domains where raw model predictions tend to exhibit high error rates.
\begin{table}[t]
\centering
\caption{ECAM: Comparison of proposed normalizers with baselines.}
\resizebox{\linewidth}{!}{
\begin{tabular}{|c|c|c|c|c|c|c|}
\hline
Normalizer  & Tiny & Base & Small & Medium & Turbo & Large \\ \hline 
No-norm  & 105.49 & 106.49 & 97.04 & 82.81 & 84.02 & 74.32 \\  
Basic & 96.44 & 94.62 & 79.79 & 67.91 & 65.82 & 50.27 \\ 
Number & 98.18 & 98.38 & 86.13 & 73.74 & 77.31 & 59.25 \\  
English & 94.57 & 94.37 & 79.86 & 67.47 & 65.39 & 50.07 \\ 
Proposed I & 94.78 & 94.85 & 79.60 & 64.73 & 65.90 & 50.58 \\
Proposed II  & 94.75 & 94.44 & 79.64 & 67.02 & 65.14 & 49.48 \\
Proposed III & 94.65 & 94.58 & 79.33 & 64.18 & 65.24 & 50.15 \\   \hline
\end{tabular}}
\label{table:normalizers_ecam}
\end{table}
\begin{table}[t]
\centering
\caption{FORDEC: Comparison of proposed normalizers with baselines.}
\resizebox{\linewidth}{!}{
\begin{tabular}{|c|c|c|c|c|c|c|}
\hline
Normalizer  & Tiny & Base & Small & Medium & Turbo & Large \\ \hline 
No-norm  & 118.98 & 96.48 & 77.56 & 69.42 & 64.20 & 63.59 \\  
Basic & 104.85 & 84.42 & 61.19 & 54.51 & 46.46 & 43.38 \\ 
Number & 107.37 & 88.66 & 68.74 & 61.66 & 55.67 & 52.68 \\  
English & 104.28 & 81.39 & 60.90 & 54.34 & 46.06 & 42.69 \\ 
Proposed I & 103.33 & 83.22 & 61.26 & 54.48 & 46.32 & 43.62 \\
Proposed II  & 104.73 & 81.66 & 61.28 & 54.49 & 45.96 & 43.34 \\
Proposed III & 103.32 & 80.05 & 61.09 & 54.38 & 45.96 & 43.09 \\   \hline
\end{tabular}}
\label{table:normalizers_fordec}
\end{table}
\begin{table}[t]
\centering
\caption{Interview: Comparison of proposed normalizers with baselines.}
\resizebox{\linewidth}{!}{
\begin{tabular}{|c|c|c|c|c|c|c|}
\hline
Normalizer  & Tiny & Base & Small & Medium & Turbo & Large \\ \hline 
No-norm  & 68.26 & 51.21 & 45.08 & 45.64 & 34.10 & 34.10 \\  
Basic & 59.26 & 41.49 & 34.95 & 37.49 & 25.18 & 23.82 \\ 
Number & 66.79 & 48.93 & 42.73 & 43.98 & 32.01 & 31.08 \\  
English & 58.96 & 41.53 & 34.92 & 37.46 & 25.12 & 23.82 \\ 
Proposed I & 59.72 & 41.50 & 35.35 & 37.58 & 25.38 & 24.13 \\
Proposed II  & 59.35 & 41.41 & 34.78 & 37.44 & 25.05 & 23.75 \\
Proposed III & 59.37 & 41.56 & 35.29 & 37.58 & 25.33 & 24.07 \\   \hline
\end{tabular}}
\label{table:normalizers_interview}
\end{table}
\begin{table}[t]
\centering
\caption{Landing: Comparison of proposed normalizers with baselines.}
\resizebox{\linewidth}{!}{
\begin{tabular}{|c|c|c|c|c|c|c|}
\hline
Normalizer  & Tiny & Base & Small & Medium & Turbo & Large \\ \hline 
No-norm  & 95.86 & 140.83 & 98.66 & 84.34 & 86.30 & 82.10 \\  
Basic & 90.78 & 118.61 & 86.56 & 67.72 & 80.12 & 65.59 \\ 
Number & 90.65 & 123.06 & 90.42 & 73.47 & 81.49 & 69.37 \\  
English & 87.43 & 119.16 & 86.18 & 67.64 & 75.40 &  65.48\\ 
Proposed I & 87.73 & 118.93 & 86.36 & 66.19 & 76.14 & 65.10 \\
Proposed II  & 87.67 & 119.25 & 85.74 & 67.61 & 75.08 & 64.86 \\
Proposed III & 87.72 & 119.40 & 85.92 & 66.83 & 75.22 & 64.58 \\   \hline
\end{tabular}}
\label{table:normalizers_landing}
\end{table}
\begin{table}[t]
\centering
\caption{Takeoff: Comparison of proposed normalizers with baselines.}
\resizebox{\linewidth}{!}{
\begin{tabular}{|c|c|c|c|c|c|c|}
\hline
Normalizer  & Tiny & Base & Small & Medium & Turbo & Large \\ \hline 
No-norm  & 119.52 & 121.93 & 113.88 & 94.72 & 85.41 & 77.44 \\  
Basic & 123.36 & 110.78 & 93.39 & 60.46 & 60.67 & 46.15 \\ 
Number & 107.50 & 109.88 & 98.12 & 71.71 & 71.14 & 55.41 \\  
English & 112.02 & 103.74 & 93.63 & 60.11 & 60.11 & 45.69 \\ 
Proposed I & 105.99 & 104.62 & 90.54 & 60.50 & 60.79 & 46.14 \\
Proposed II  & 112.04 & 103.75 & 93.62 & 59.49 & 59.77 & 45.68 \\
Proposed III & 108.51 & 103.69 & 90.64 & 59.79 & 59.64 & 44.89 \\   \hline
\end{tabular}}
\label{table:normalizers_takeoff}
\end{table}

\subsection{Effect of Finetuning}
 Table~\ref{table:lora_finetune_details} provides details about LoRA fine-tuning for different sizes of Whisper models. It presents the total number of parameters in each model, the number of additional LoRA parameters introduced during fine-tuning, and the percentage of LoRA parameters relative to the total model size. LoRA requires only a small fraction (0.8\% to 1.6\%) of the total model parameters, reducing the number of trainable parameters while still allowing effective adaptation.\\
 \indent LoRA fine-tuning on Whisper Large to Whisper Tiny models with various learning rates is given in Tables~\ref{table:lora_large}~to~\ref{table:lora_tiny}. The fine-tuning results across Whisper models of varying sizes (Tiny, Base, Small, and Medium) demonstrate that LoRA fine-tuning leads to significant reductions in WER across all configurations, with the extent of improvement depending on model size, normalization technique, and learning rate. The pre-trained models exhibit relatively high WER, particularly in the absence of normalization, with the No-norm baseline consistently yielding the worst performance. Fine-tuning improves recognition accuracy substantially, with Proposed II and English normalizers achieving the lowest WER across most scenarios.\\
\indent For Whisper Medium and Small models, the optimal learning rate appears to be 1e-3, where Proposed II and English yield the lowest WER (32.67\% and 32.97\% for Medium; 39.18\% and 39.11\% for Small).
However, for Whisper Base and Tiny models, higher learning rates (1e-3) occasionally lead to performance degradation before normalization. Notably, the No-norm baseline for Whisper Tiny at 1e-3 results in a WER of 96.31\%, exceeding that of the pre-trained model, while the normalized WER being lower for finetuned model over pre-trained.
Among the normalization techniques, Proposed II and English consistently outperform other approaches, demonstrating their effectiveness in improving ASR accuracy post-fine-tuning.
\begin{table}[t]
\centering
\caption{LoRA Finetuning details}
\resizebox{\linewidth}{!}{
\begin{tabular}{|c|c|c|c|}
\hline
Model & Total parameters  &  LoRA parameters & Percentage (\%)  \\ \hline 
Tiny & $38,350,464$ & $589,824$ & $1.5380$
\\
Base & $73,773,568$
& $1,179,648$
& $1.5990$
\\
Small & $245,273,856$
& $3,538,944$
& $1.4429$
\\
Medium & $773,295,104$
& $9,437,184$
 &$1.2204$
 \\
Turbo & $815,431,680$
& $6,553,600$
& $0.8037$
\\
Large & $1,559,219,200$
& $15,728,640$
& $1.009$
 \\ \hline
\end{tabular}}
\label{table:lora_finetune_details}
\end{table}
\begin{table}[t]
\centering
\caption{LoRA fine-tuning on Whisper Large model with various learning rates. The numbers indicate WER in \%.}
\resizebox{\linewidth}{!}{
\begin{tabular}{|c|c|c|c|c|}
\hline
Normalizer  & pre-trained & lr=1e-5 & lr=1e-4 & lr=1e-3  \\ \hline No-norm  & 68.49 & 58.83 & 64.36 & 55.65 \\  
Basic  & 52.23 & 27.96 & 37.09 & 27.37 \\ 
Number  & 59.76 & 46.10 & 48.30 & 50.08 \\ 
English  & 52.08 & 27.80 & 36.71 & 26.36 \\ 
Proposed I  & 52.74 & 32.72 & 37.35 & 38.37 \\ 
Proposed II  & 52.00 & 27.65 & 36.41 & 26.26 \\ 
Proposed III  & 52.41 & 28.24 & 36.60 & 27.00 \\    \hline
\end{tabular}}
\label{table:lora_large}
\end{table}
\begin{table}[t]
\centering
\caption{LoRA fine-tuning on Whisper Turbo model with various learning rates. The numbers indicate WER in \%.}
\resizebox{\linewidth}{!}{
\begin{tabular}{|c|c|c|c|c|}
\hline
Normalizer  & pre-trained & lr=1e-5 & lr=1e-4 & lr=1e-3  \\ \hline No-norm  & 70.20 & 61.82 & 64.67 & 65.06 \\  
Basic  & 49.49 & 28.18 & 29.04 & 31.02 \\  
Number  & 62.18 & 43.08 & 46.64 & 47.61 \\  
English  & 48.88 & 28.01 & 28.81 & 30.40 \\  
Proposed I  & 49.68 & 29.17 & 29.98 & 31.70 \\  
Proposed II  & 48.69 & 28.24 & 28.88 & 30.32 \\   
Proposed III  & 48.71 & 28.40 & 28.67 & 30.55 \\      \hline
\end{tabular}}
\label{table:lora_turbo}
\end{table}
\begin{table}[t]
\centering
\caption{LoRA fine-tuning on Whisper Medium model with various learning rates. The numbers indicate WER in \%.}
\resizebox{\linewidth}{!}{
\begin{tabular}{|c|c|c|c|c|}
\hline
Normalizer  & pre-trained & lr=1e-5 & lr=1e-4 & lr=1e-3  \\ \hline No-norm  & 81.64& 60.10 &66.84 &63.85 \\  
Basic  & 62.96 & 35.69 &36.12 & 33.22\\    
Number  & 70.84 & 50.35 &50.46 & 49.27\\    
English  & 62.43 & 34.48& 35.96& 32.97\\   
Proposed I  & 63.19 &36.87 &36.76 & 35.63\\   
Proposed II  & 62.20 &34.60 & 35.18&32.67 \\    
Proposed III  & 62.54 &34.24 & 36.28& 33.22\\      \hline
\end{tabular}}
\label{table:lora_medium}
\end{table}
\begin{table}[t]
\centering
\caption{LoRA fine-tuning on Whisper Small model with various learning rates. The numbers indicate WER in \%.}
\resizebox{\linewidth}{!}{
\begin{tabular}{|c|c|c|c|c|}
\hline
Normalizer & pre-trained & lr=1e-5 & lr=1e-4 & lr=1e-3  \\ \hline 
No-norm  & 85.64 & 75.67&70.33 & 63.72\\  
Basic  & 69.35&  43.26&48.63 & 39.88\\   
Number  & 77.41& 57.39& 67.56& 61.53\\   
English  & 69.16& 42.74 &47.81& 39.11\\
Proposed I  &69.05 & 43.11&61.84 & 56.30\\
Proposed II  & 68.87& 42.49&47.73 & 39.18\\    
Proposed III  & 68.76& 42.39& 49.09& 40.19\\      \hline
\end{tabular}}
\label{table:lora_small}
\end{table}
\begin{table}[t]
\centering
\caption{LoRA fine-tuning on Whisper Base model with various learning rates. The numbers indicate WER in \%.}
\resizebox{\linewidth}{!}{
\begin{tabular}{|c|c|c|c|c|}
\hline
Normalizer & pre-trained & lr=1e-5 & lr=1e-4 & lr=1e-3  \\ \hline 
No-norm  & 96.00& 88.56&81.45 &73.95\\  
Basic  & 84.70&60.06 &62.64 &56.57\\  
Number & 88.70& 72.29& 69.55&72.40\\    
English  & 84.58& 59.92& 60.40&56.08\\ 
Proposed I  &83.10 & 60.49& 58.55&69.97\\ 
Proposed II  &84.25 & 60.11& 59.95&56.00\\   
Proposed III  &82.96 &60.55 & 60.24&57.23\\       \hline
\end{tabular}}
\label{table:lora_base}
\end{table}
\begin{table}[t]
\centering
\caption{LoRA fine-tuning on Whisper Tiny model with various learning rates. The numbers indicate WER in \%.}
\resizebox{\linewidth}{!}{
\begin{tabular}{|c|c|c|c|c|}
\hline
Normalizer  & pre-trained & lr=1e-5 & lr=1e-4 & lr=1e-3  \\ \hline 
No-norm  & 94.41&92.06 &86.34 &96.31\\  
Basic  & 91.73& 90.80&66.24 &75.79\\  
Number &89.13 & 86.93&76.28 &84.88\\    
English  & 88.37&84.68 & 66.33&74.49\\ 
Proposed I  &85.68 &82.94 &67.17 &74.91\\ 
Proposed II  &88.41 &84.78 &66.10 &74.69\\   
Proposed III  & 88.21& 84.99&67.04 &74.17\\       \hline
\end{tabular}}
\label{table:lora_tiny}
\end{table}

\section{Challenges with multi-lingual speech}
Table~\ref{table:translation_challenges} shows instances where the Whisper model transcriptions struggles with unexpected translation, often misinterpreting words or phrases based on phonetic similarities rather than contextual meaning. For example, "Gut" is incorrectly transcribed as "Good," reflecting a bias toward English interpretations. Similarly, longer phrases exhibit structural differences that lead to errors in word order and meaning retention.\\ 
\indent Table~\ref{table:close_calls} presents cases where the model's predictions are very similar to the reference text but still contain subtle inaccuracies. These transcription errors often involve homophones or phonetically similar words, such as "slats low" misrecognized as "sled slow" and "CAT3 single" transcribed as "cut three single." 
\begin{table}[t]
\centering
\caption{Transcription with unexpected translation}
\resizebox{\linewidth}{!}{
\begin{tabular}{|c|c|}
\hline
Reference  &  Prediction  \\ \hline 
Ist confirmed  &  That was confirmed \\
Gut & Good \\ 
Blaues system ist natürlich verloren & The blue system is of course lost \\ 
daraufhin ein spoiler-pair & then a spoiler pair \\ \hline
\end{tabular}}
\label{table:translation_challenges}
\end{table}
\begin{table}[t]
\centering
\caption{Transcription errors: Words with close phonetics.}
\resizebox{0.75\linewidth}{!}{
\begin{tabular}{|c|c|}
\hline
Reference  &  Prediction  \\ \hline 
clear flight control & okay, flight control \\
clear flight control & flight control \\
read status & wave status \\
slats low & sled low \\
CAT3 single & cut three single \\
Inop systems & In-hub systems \\ \hline
\end{tabular}}
\label{table:close_calls}
\end{table}

\end{document}